\documentclass[runningheads]{llncs}
\usepackage{graphicx}\usepackage{graphicx}
\usepackage{subfigure}
\usepackage{epstopdf}
\usepackage{enumerate}
\usepackage{amsmath}
\usepackage{amsfonts,amssymb}
\usepackage{algorithm}
\usepackage{algorithmicx}
\usepackage{algpseudocode}
\usepackage{multirow}
\usepackage{multicol}
\usepackage{array}
\newcommand{\PreserveBackslash}[1]{\let\temp=\\#1\let\\=\temp}
\newcolumntype{C}[1]{>{\PreserveBackslash\centering}p{#1}}
\newcolumntype{R}[1]{>{\PreserveBackslash\raggedleft}p{#1}}
\newcolumntype{L}[1]{>{\PreserveBackslash\raggedright}p{#1}}
\usepackage{caption}
\usepackage{amssymb}
\usepackage{bbding}

\begin{document}
\title{Multi-organ Segmentation via Co-training Weight-averaged Models from Few-organ Datasets}
\titlerunning{Multi-organ Segmentation via Co-training Weight-averaged Models}

\author{
    Rui Huang\inst{1},
    Yuanjie Zheng\inst{2(}\Envelope\inst{)},
    Zhiqiang Hu\inst{1},
    Shaoting Zhang\inst{1}, \\
    \and Hongsheng Li\inst{3(}\Envelope\inst{)}
}
\authorrunning{Rui Huang et al.}

\institute{SenseTime Research, \email{\{huangrui@sensetime.com\}}
\and School of Information Science and Engineering at Shandong Normal University \\ $^3$ CUHK-SenseTime Joint Laboratory, The Chinese University of Hong Kong, \email{\{hsli@ee.cuhk.edu.hk\}}\\ 
}

\maketitle              
\begin{abstract}
Multi-organ segmentation requires to segment multiple organs of interest from each image.
However, it is generally quite difficult to collect full annotations of all the organs on the same images, as some medical centers might only annotate a portion of the organs due to their own clinical practice. In most scenarios, one might obtain annotations of a single or a few organs from one training set, and obtain annotations of the other organs from another set of training images. 
Existing approaches mostly train and deploy a single model for each subset of organs, which are memory intensive and also time inefficient. In this paper, we propose to co-train weight-averaged models for learning a unified multi-organ segmentation network from few-organ datasets. Specifically, we collaboratively train two networks and let the coupled networks teach each other on un-annotated organs. To alleviate the noisy teaching supervisions between the networks, the weighted-averaged models are adopted to produce more reliable soft labels. In addition, a novel region mask is utilized to selectively apply the consistent constraint on the un-annotated organ regions that require collaborative teaching, which further boosts the performance. Extensive experiments on three publicly available single-organ datasets LiTS \cite{bilic2019liver}, KiTS \cite{heller2019kits19}, Pancreas \cite{simpson2019large} and manually-constructed single-organ datasets from MOBA \cite{gibson2018automatic} show that our method can better utilize the few-organ datasets and achieves superior performance with less inference computational cost.


\end{abstract}
\section{Introduction}

\begin{figure}[tp]
    \begin{minipage}[b]{1\linewidth}
      \centering
      {\includegraphics[width=10cm]{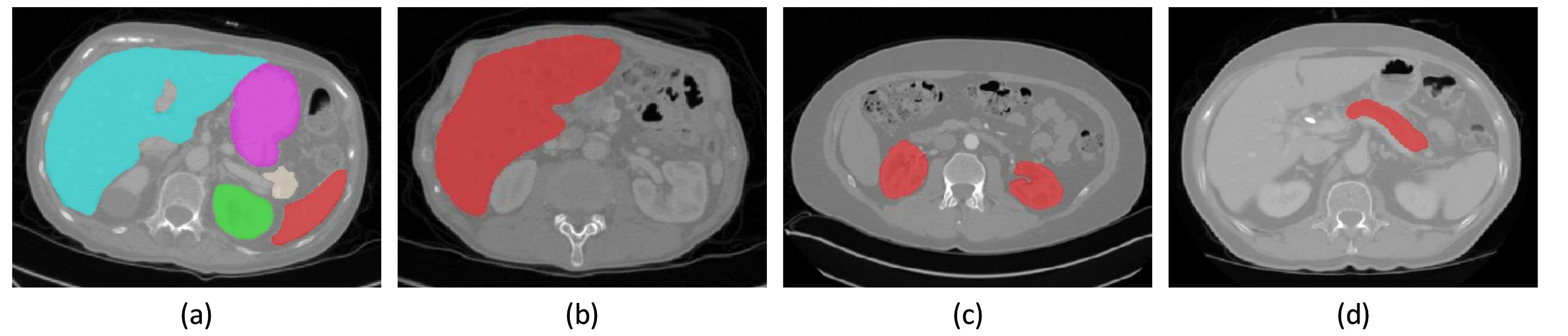}}
    \end{minipage}
    \caption{(a) Multi-organ segmentation is required to achieve more comprehensive computer-aided analysis. 
    (b) LiTS dataset \cite{bilic2019liver} contains only liver annotations. (c) KiTS \cite{heller2019kits19} dataset contains only kidney annotations. (d) Pancreas \cite{simpson2019large} dataset contains only pancreas annotations.}
    \label{fig1}
\end{figure}

In medical image segmentation, obtaining multi-organ annotations on the same set of images is labor-intensive and time-consuming, where only experienced radiologists are qualified for the annotation job. On the other hand, different medical centers or research institutes might have annotated a subset of organs for their own clinical and research purposes. For instance, there are publicly available single-organ datasets, such as LiTS \cite{bilic2019liver}, KiTS \cite{heller2019kits19} and Pancreas \cite{simpson2019large}, each of which only provides a single organ's annotations, 
as shown in Fig \ref{fig1}. However, existing methods cannot effectively train a multi-organ segmentation network based on those single-organ datasets with different images.

This work focuses on learning multi-organ segmentation from few-organ datasets for abdominal computed tomography (CT) scans. An intuitive solution is to segment each organ by a separate model using a training dataset. However, this solution is computationally expensive and the spatial relationships between different organs cannot be well exploited. Furthermore, some researches adopt self-training \cite{papandreou2015weakly}, which generates pseudo labels for un-annotated organs in each dataset using a trained single-organ segmentation model, and constructs a pseudo multi-organ dataset. The multi-organ segmentation model can be learned from the pseudo multi-organ dataset. Obviously, the pseudo labels might contain much noise due to the generalization inability of each single-organ segmentation model, as well as the domain gap between different datasets. The inaccurate pseudo labels would harm the training process and limit the performance of self-training.

To tackle the challenge, we propose to co-train a pair of weight-averaged models for unified multi-organ segmentation from few-organ datasets. Specifically, to provide supervisions for un-annotated organs, we adopt the temporally weight-averaged model to generate soft pseudo labels on un-annotated organs. In order to constrain error amplification, two models' weight-averaged versions are used to provide supervisions for training each other on the un-annotated organs via consistency constraints, in a collaborative manner. A novel region mask is proposed to adaptively constrain the network to mostly utilize the soft pseudo labels on the regions of un-annotated organs. Note that our proposed framework with two networks is only adopted during training stage and only one network is used for inference without additional computational or memory overhead.

The contributions of our works are threefold: (1) We propose to co-train collaborative weight-averaged models for achieving unified multi-organ segmentation from few-organ datasets. (2) The co-training strategy, weight-averaged model and the region mask are developed for more reliable consistency training. (3) The experiment results show that our framework better utilizes the few-organ datasets and achieves superior performance with less computational cost.

\section{Related work}

Recently, CNNs have made tremendous progress for semantic segmentation. Plenty of predominant approaches have been proposed, such as DeepLab \cite{chen2018encoder}, PSPNet \cite{zhao2017pyramid} for natural images and UNet \cite{ronneberger2015u}, VoxResNet \cite{chen2018voxresnet} for medical images. Due to the difficulty of obtaining multi-organ datasets, many approaches are dedicated to the segmentation of one particular organ. Chen et al. \cite{chen2019harnessing} proposed a two-stage framework for accurate pancreas segmentation. As these approaches are designed under fully-supervised setting, they cannot be directly applied to train a multi-organ segmentation model from few-organ datasets.

Konstantin et al. \cite{dmitriev2019learning} firstly present a conditional CNN framework for multi-class segmentation and demonstrate the possibility of producing multi-class segmentations using a single model trained on single-class datasets. However, the inference time of their method is proportion to the number of organs, 
which is inefficient. Zhou et al. \cite{zhou2019prior} incorporated domain-specific knowledge for multi-organ segmentation using partially annotated datasets. But their training objective is difficult to optimized and it needs some specific optimization methods. 

Teacher-student model is a widely used technique in semi-supervised learning (SSL) and model distillation. The key idea is to transfer knowledge from a teacher to a student network via consistency training. Deep mutual learning \cite{zhang2018deep} proposed to train two networks collaboratively with the supervision from each other. Mean-teacher model \cite{tarvainen2017mean} averaged model weights over different training iterations to produce supervisions for unlabeled data. Ge et al. \cite{ge2020mutual} proposed a framework called Mutual Mean Teaching for pseudo label refinery in person re-ID. Note that these methods are designed under fully-supervised or semi-supervised settings. In this work, we exploit the integration of co-training strategy and weight-averaged models for unifying multi-organ segmentation from few-organ datasets.

\section{Method}
Single-organ datasets are special cases of few-organ datasets. Without loss of generality, we discuss how to train multi-organ segmentation networks from single-organ datasets in this section. The method can be easily extended to handle few-organ datasets. Formally, given $K$ single-organ datasets $\left\{\mathbb{D}_{1}, \cdots,\mathbb{D}_{K}\right\}$, where $\mathbb{D}_{k}=\{(x_i^k,$ $y_i^{k})|  i = 1, \cdots, N_k, k = 1,\cdots, K\}$, let $x_i^k$ and $y_i^{k}$ denote the $i$-th training sample in the $k$-th single-organ dataset and its associated binary segmentation mask for organ $k$ out of all $K$ organs. Our goal is to train a unified network that can output segmentation maps for all $K$ organs simultaneously.

\begin{figure}[t]
    \begin{center}
    {
    {\includegraphics[width=1\linewidth]{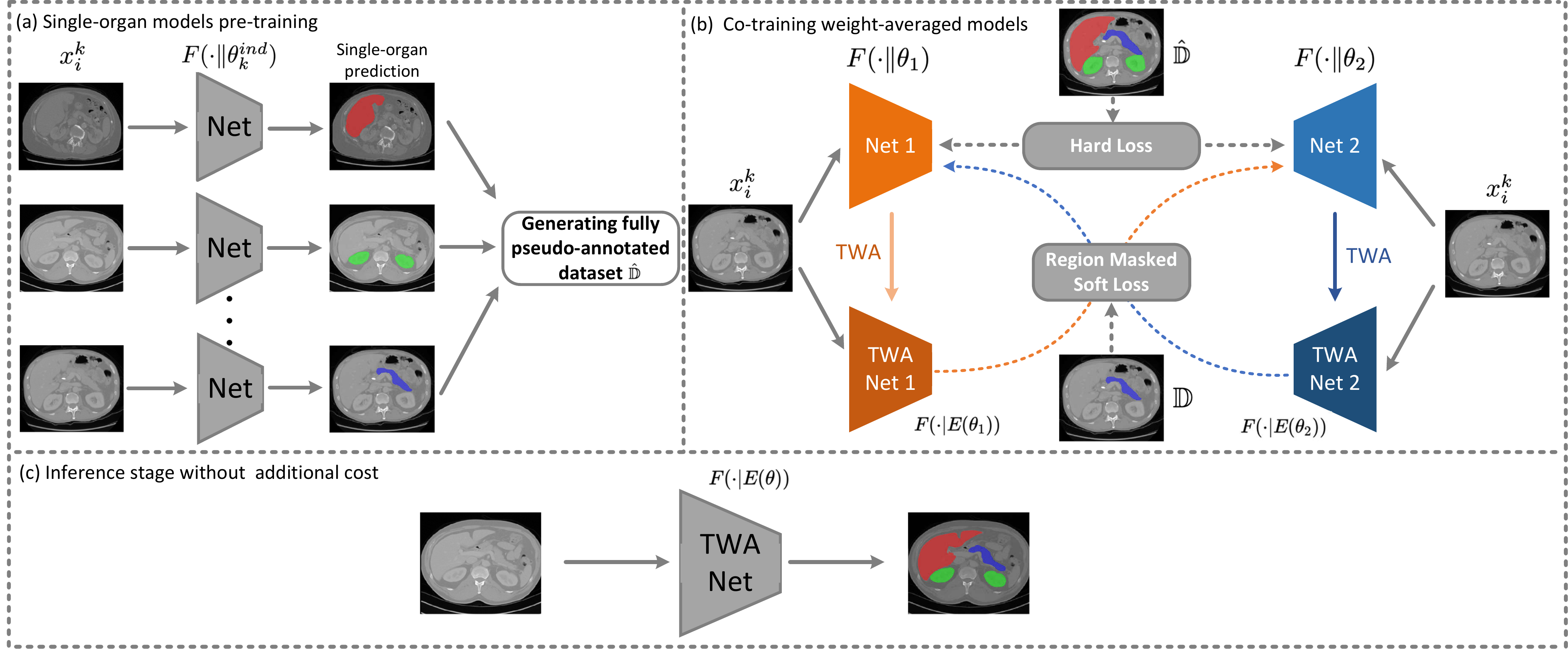}}}
    \caption{(a) The pipeline of generating fully pseudo-annotated dataset. (b) The overall framework of our method. 
    (c) In inference phase, only one network is used without requiring additional computational cost.}
    \label{framework}
    \end{center}
\end{figure}

{\bf Pre-training single-organ and multi-organ models.} We choose DeepLab \cite{chen2018encoder} with dilated Resnet-50 \cite{zhang2018context} and IBN modules \cite{pan2018two} as our segmentation backbone for its strong capability on different semantic segmentation tasks. Note that other segmentation networks could also be adopted in our proposed framework. We first pre-train $K$ segmentation models on the $K$ single-organ datasets, respectively. Each model is responsible for segmentation of one individual organ, denoted as $F(\cdot\|\theta_k^{ind})$, where $\theta_k^{ind}$ denotes network parameters of the $k$-th single-organ network. With the pre-trained models, for each single-organ dataset $\mathbb{D}_k$, we can generate pseudo labels for those un-annotated organs to create a fully-annotated dataset with pseudo labels $\hat{\mathbb{D}}_k$, where the label of organ $k$ is manually annotated and the others are hard pseudo labels(see Fig \ref{framework}(a)). We can then construct a joint fully pseudo-annotated dataset $\hat{\mathbb{D}} = \{\hat{\mathbb{D}}_{1},\cdots,\hat{\mathbb{D}}_{K}\}$.

Based on pseudo segmentation masks, we can pre-train a multi-organ segmentation model $\hat{F}$ on $\hat{\mathbb{D}}$. Obviously, the quality of pseudo labels is vital to the final performance. It is inevitable that some pseudo label maps might be inaccurate due to the generalization inability of each single-organ segmentation model.
The noisy pseudo labels would therefore harm the final segmentation accuracy. 

{\bf Co-training weight-averaged models for pseudo label regularization.} Our framework is illustrated in Fig \ref{framework}(b). 
We train a pair of collaborative networks, $F(\cdot\|\theta_1)$ and $F(\cdot\|\theta_2)$, with the same structure as our pre-trained multi-organ network $\hat{F}$ but with randomly initialized parameters. The training utilizes both the fully pseudo-annotated dataset $\hat{\mathbb{D}}$ with hard pseudo labels and the original dataset $\mathbb{D} = \{\mathbb{D}_1, \cdots, \mathbb{D}_K\}$ with online generated soft pseudo labels. 
Both the networks are trained with the weighted focal loss and dice loss using hard labels in the fully pseudo-annotated dataset $\hat{\mathbb{D}}$, for handling the high variability of organ size in abdomen. The weighted focal loss is defined as
\begin{equation}
    \mathcal{L}_\mathrm{focal}(\theta) = -\sum_{k=1}^K\sum_{i=1}^{N_k}\sum_{c=1}^{C}\alpha_{c}\left(1-F(x_i^k\lvert\theta)_c\right)^{\gamma} \log \left(F(x_i^k\lvert\theta)_c\right),
\end{equation}
where $c$ denotes the $c$-th organ class and $F(x_i^k\lvert\theta)_c$ is model's estimated probability that a pixel is correctly classified. $\alpha_c$ is the weight of each organ $c$, which is inversely proportional to each organ's average size. The parameter $\gamma$ is set as 2 empirically.
The dice loss can be formulated as
\begin{equation}
    \mathcal{L}_\mathrm{dice}(\theta) = \sum_{k=1}^K\sum_{i=1}^{N_k}\sum_{c=1}^{C}\left(1-2 \frac{\sum \hat{y}_{i}^{k,c} F(x_i^k\lvert\theta)_c + \epsilon}{\sum \hat{y}_{i}^{k,c}+\sum F(x_i^k\lvert\theta)_c+\epsilon}\right),
\end{equation}
where $\hat{y}_{i}^{k}$ and $F(x_i^k\lvert \theta)_c$ represent the hard labels and model's predictions for organ $c$, respectively. $\epsilon$ is a small value to ensure numerical stability. Note that the above losses are applied to both networks' parameters, $\theta_1$ and $\theta_2$.

Since the hard pseudo labels are quite noisy, to properly regularize the learning process, we also adopt the online generated soft pseudo labels for un-annotated organs when training the networks on the original data $\mathbb{D}$. For training network 1, $F(\cdot\|\theta_1^{(t)})$, at iteration $t$ with image $x_i^k\in \mathbb{D}_k$, the ground-truth labels for organ $k$ are used while other organs' predicted soft labels are generated from the network 2, $F(\cdot\|\mathbb{E}_t(\theta_2))$, with temporally averaged parameters $\mathbb{E}_t(\theta_2)$:
\begin{align}
	\mathbb{E}_t(\theta_2) = \alpha\mathbb{E}(\theta_2^{(t-1)}) + (1-\alpha) \theta_2^{(t)},
\end{align}
where $\alpha \in [0,1]$ controls how fast the parameters are temporally averaged. Similarly, network 2's parameters $\theta_2^{(t)}$ are trained by temporally weight-averaged model of network 1 $F(\cdot\|\mathbb{E}_t(\theta_1))$'s predictions. Intuitively, the temporally weight-averaged version is a temporal ensemble of a network over its past iterations, which can generate more robust online soft pseudo labels for the un-annotated organs than the network at a specific iteration. In addition, we adopt one network's temporal average to supervise the other network. This strategy can avoid each network using its own previous iterations' predictions as supervisions, which might amplify its segmentation errors from previous iterations. 

For each image $x_i^k$, pixels belong to organ-$k$ are with ground-truth annotations. We would avoid adopting the soft pseudo labels for training networks on regions of ground-truth organs. We morphologically dilate the hard pseudo labels for each un-annoatated organ to generate a region mask ${\cal T}(y_i^k)$:
\[ \mathcal{T}(y_i^k) = 
\begin{cases}
1, & \text{regions without annotations or background}, \\
0, & \text{regions with organ-$k$ ground-truth annotations}.
\end{cases}
\]
Therefore, the segmentation loss with soft pseudo labels are formulated as:
\begin{gather}
    \mathcal{L}_\mathrm{soft}(\theta_1^{(t)}\lvert\theta_2^{(t)}) = -\sum_{k=1}^K\sum_{i=1}^{N_k}(\mathcal{T}(y_i^{k})\cdot F(x_i^k\lvert \mathbb{E}_t(\theta_2) \cdot \log F(x_i^k\lvert\theta_1^{(t)})), \\ 
    \mathcal{L}_\mathrm{soft}(\theta_2^{(t)}\lvert\theta_1^{(t)}) = -\sum_{k=1}^K\sum_{i=1}^{N_k}(\mathcal{T}(y_i^{k})\cdot F(x_i^k\lvert \mathbb{E}_t(\theta_1) \cdot \log F(x_i^k\lvert\theta_2^{(t)})).
\end{gather}

The key difference between our method and mean teacher \cite{tarvainen2017mean} is that we use the temporally weight-averaged version of one network to supervise another network. Such a collaborative training manner can further decouple the networks' predictions. In addition, the region masks are important to enforce the soft label supervisions are only applied to un-annotated regions.

{\bf Overall segmentation loss.} Our framework is trained with the supervision of the hard loss and the soft loss. The overall loss function optimizes the two networks simultaneously, which is formulated as:
\begin{align}
    \mathcal{L}(\theta_1,\theta_2) = & \lambda_\mathrm{focal}(\mathcal{L}_\mathrm{focal}(\theta_1)+\mathcal{L}_\mathrm{focal}(\theta_2)) + \lambda_\mathrm{dice}(\mathcal{L}_\mathrm{dice}(\theta_1)+\mathcal{L}_\mathrm{dice}(\theta_2)) \notag \\ 
    + & \lambda_\mathrm{rampup}\lambda_\mathrm{soft}(\mathcal{L}_\mathrm{soft}(\theta_1\lvert\theta_2)+\mathcal{L}_\mathrm{soft}(\theta_2\lvert\theta_1)),
\end{align}
where $\lambda_\mathrm{focal}, \lambda_\mathrm{dice}$ and $\lambda_\mathrm{soft}$ are loss weights. Since the predictions at early training stages might not be accurate, we apply a ramp-up strategy to gradually increase $\lambda_\mathrm{rampup}$, which makes the training process more stable.

\section{Experiments}
The proposed framework was evaluated on three publicly available single-organ datasets, LiTS \cite{bilic2019liver}, KiTS \cite{heller2019kits19}, Pancreas \cite{simpson2019large} and a manually-constructed single-organ dataset MOBA \cite{gibson2018automatic}. LiTS consists of 131 training and 70 test CT scans with liver annotations, provided by several clinical sites. 
KiTS consists of 210 training and 90 test CT scans with kidney annotations, collected from 300 patients who underwent partial or radical nephrectomy. Pancreas consists of 281 training and 139 test CT scans with pancreas annotations, provided by Memorial Sloan Kettering Cancer Center. Since the annotation is only available for the training set, we use their training sets in our experiments. MOBA is a multi-organ dataset with 90 CT scans drawn from two clinical sites. 
The authors of \cite{gibson2018automatic} provided segmentation masks of eight organs, including spleen, left kidney, gallbladder, esophagus, liver, stomach, pancreas and duodenum. 
Specifically, the multi-organ segmentation masks are binarized and stored separately, i.e., we have manually constructed eight single-organ datasets. All datasets are divided into training and test sets with a 4:1 ratio. We use the Dice-Score-Coefficient (DSC) and Hausdorff Distance (HD) as the evaluation metric: $\operatorname{DSC}(\mathcal{P}, \mathcal{G})=\frac{2 \times|\mathcal{P} \cap \mathcal{G}|}{|\mathcal{P}|+|\mathcal{G}|}$, where $\mathcal{P}$ is the binary prediction and $\mathcal{G}$ is the ground truth. HD measures the largest distance from points in $\mathcal{P}$ to its nearest neighbour in $\mathcal{G}$ and the distances of two directions are averaged to get the final metric: $\operatorname{HD}(\mathcal{P}, \mathcal{G})=(d_H(\mathcal{P}, \mathcal{G})+d_H(\mathcal{G}, \mathcal{P}))/2$.

For preprocessing, all the CT scans are re-sampled to $1\times 1\times 3$ mm. The CT intensity values are re-scaled to [0, 1] using a window of [-125, 275] HU for better contrast. We then center crop a $352\times 352$ patch as the network input.

\subsection{Implementation details.}
All models were trained for 10 epochs using synchronized SGD on 8 NVIDIA 1080 Ti GPUs with a minibatch of 24 (3 images per GPU). The initial learning rate is 0.05 and a $cosine$ learning rate policy is employed. Weight decay of 0.0005 and momentum of 0.9 are used during training. The hyper-parameters $\lambda_\mathrm{focal}, \lambda_\mathrm{dice}$ and $\lambda_\mathrm{soft}$ are set to 1.0, 0.1 and 0.1, respectively. The smoothing coefficient $\alpha$ is set as 0.999. During inference, only one of the two weight-averaged models with better validation performance is used as the final model. 

\subsection{Experiments on LiTS-KiTS-Pancreas dataset}
For the single-organ datasets, LiTS, KiTS and Pancreas, we first train three single-organ models separately for each organ, denoted as ``individual'' in Table \ref{tab:1}. It achieves 95.90\%, 95.30\%, 77.05\% DSC for liver, kidney, pancreas, respectively, and an average DSC of 89.41\%. 


\begin{table}[!t]
    \tiny
    \minipage[t]{0.62\textwidth}
    \centering
    \captionsetup{font={scriptsize}}
    \caption{Ablation studies of our proposed methods on the LiTS-KiTS-Pancreas dataset. CT: co-training strategy. WA: weight-averaged model. RM: region mask.}
    \label{tab:1}
    \begin{tabular}{C{2.5cm}C{1cm}C{1cm}C{1cm}C{1cm}}
        \hline
        Method & Liver & Kidney & Pancreas & Avg DSC \\
        \hline
        individual & 95.90 & \bf{95.30} & 77.05 & 89.41  \\
        self-training & 95.94 & 94.02 & 78.54 & 89.50  \\
        CT & 95.89 & 94.42 & 78.15 & 89.49  \\
        WA & 95.90 & 94.49 & 78.52 & 89.63  \\
        CT+WA & 95.93 & 94.31 & 79.12 & 89.78  \\
        CT+WA+RM & \bf{95.96} & 95.01 & 79.25 & 90.07  \\
        Ours(CT+WA+RM+IBN) & 95.90 & 94.98 & \bf{79.78} & \bf{90.22}  \\
        \hline
    \end{tabular}
    \endminipage\hspace{5pt}
    \minipage[t]{0.35\textwidth}
    \centering
    \captionsetup{font={scriptsize}}
    \caption{DSC(\%) and execution time of conditionCNN \cite{dmitriev2019learning} and our method.}
    \label{tab:2}
    \renewcommand\arraystretch{1.32}
    \begin{tabular}{p{1.5cm}p{2cm}p{1cm}}
        \hline
        Method & conditionCNN \cite{dmitriev2019learning} & Ours \\
        \hline
        Liver & \bf{95.93} & 95.90  \\
        Kidney & \bf{95.33} & 94.98 \\
        Pancreas & 77.90 & \bf{79.78} \\
        \hline
        Avg DSC & 89.72 & \bf{90.22} \\
        Time (s) & 12.9 & \bf{4.28} \\
        \hline
    \end{tabular}
    \endminipage
\end{table}

{\bf Ablation study.} In this section, we evaluate each component's effect in our framework. The ablation results are shown in Table \ref{tab:1}. We can observe that when training a multi-organ segmentation model directly with hard pseudo labels (denoted as ``self-training''), the performance is slightly better than single-organ models (89.41\% to 89.50\%), which means that even the noisy pseudo labels can improve the segmentation of un-annotated organs. Meanwhile, by applying the co-training scheme, weight-averaged model, region mask and IBN module, our proposed framework achieves a remarkable improvement of 0.81\% in terms of average DSC (89.41\% to 90.22\%) without any additional computational cost for inference. Especially, we observe a significant performance gain of 2.73\% for the segmentation of the pancrea, which is more challenging because of its smaller sizes and irregular shapes. Note that when only applying the co-training scheme, the performance is just comparable with self-training, which demonstrates that using the weight-averaged model for supervising the other model can produce more reliable soft labels. The weight-averaged model, region mask and IBN module bring performance gains of 0.29\%, 0.29\%, and 0.15\%, respectively.


\begin{table}[!t]
    \vspace{-15pt}
    \tiny
    \renewcommand\arraystretch{1}
    \centering
    \captionsetup{font={scriptsize}}
    \caption{DSC(\%) and HD(mm) comparison on MOBA dataset.}
    \label{tab:3}
    \begin{tabular}{C{1.5cm}|C{1cm}C{1cm}C{1.5cm}C{1cm}C{2cm}|C{1cm}C{1.5cm}C{1cm}}
    \hline
    & \multicolumn{5}{c|}{DSC(\%)} & \multicolumn{3}{c}{HD(mm)} \\
    \hline
    Organ & individual & combine & self-training & Ours & conditionCNN \cite{dmitriev2019learning} & individual & self-training & Ours \\
    \hline
    Spleen & \bf{96}.00 & 95.28 & 95.69 & 94.95 & 80.77 & \bf{16.81} & 32.54 & 23.15 \\
    Kidney(L) & 94.51 & 94.35 & 94.60 & \bf{95.03} & 81.51 & 29.20 & 26.29 & \bf{22.33} \\
    Gallbladder & 78.59 & 79.55 & \bf{80.43} & 79.65 & 66.15 & 54.22 & \bf{31.58} & 34.35 \\
    Esophagus & 66.07 & 62.90 & 71.87 & \bf{72.25} & 55.03 & 26.85 & 26.58 & \bf{24.47} \\
    Liver & \bf{96.61} & 96.12 & 96.23 & 96.18 & 94.47 & \bf{31.99} & 45.89 & 43.07 \\
    Stomach & \bf{91.35} & 87.65 & 90.08 & 89.68 & 82.56 & 57.64 & 43.71 & \bf{42.93} \\
    Pancreas & 78.04 & 73.69 & 78.10 & \bf{79.35} & 60.55 & 28.61 & 31.04 & \bf{29.68} \\
    duodenum & 58.16 & 54.53 & 57.72 & \bf{61.63} & 47.60 & 41.28 & 38.43 & \bf{35.08} \\
    \hline
    Avg & 82.41 & 82.02 & 83.09 & \bf{83.60} & 62.37 & 35.82 & 34.51 & \bf{31.88} \\
    \hline
    \end{tabular}
\end{table}

\begin{figure}[!t]
    \begin{center}
    {
    {\includegraphics[width=1\linewidth, height=5cm]{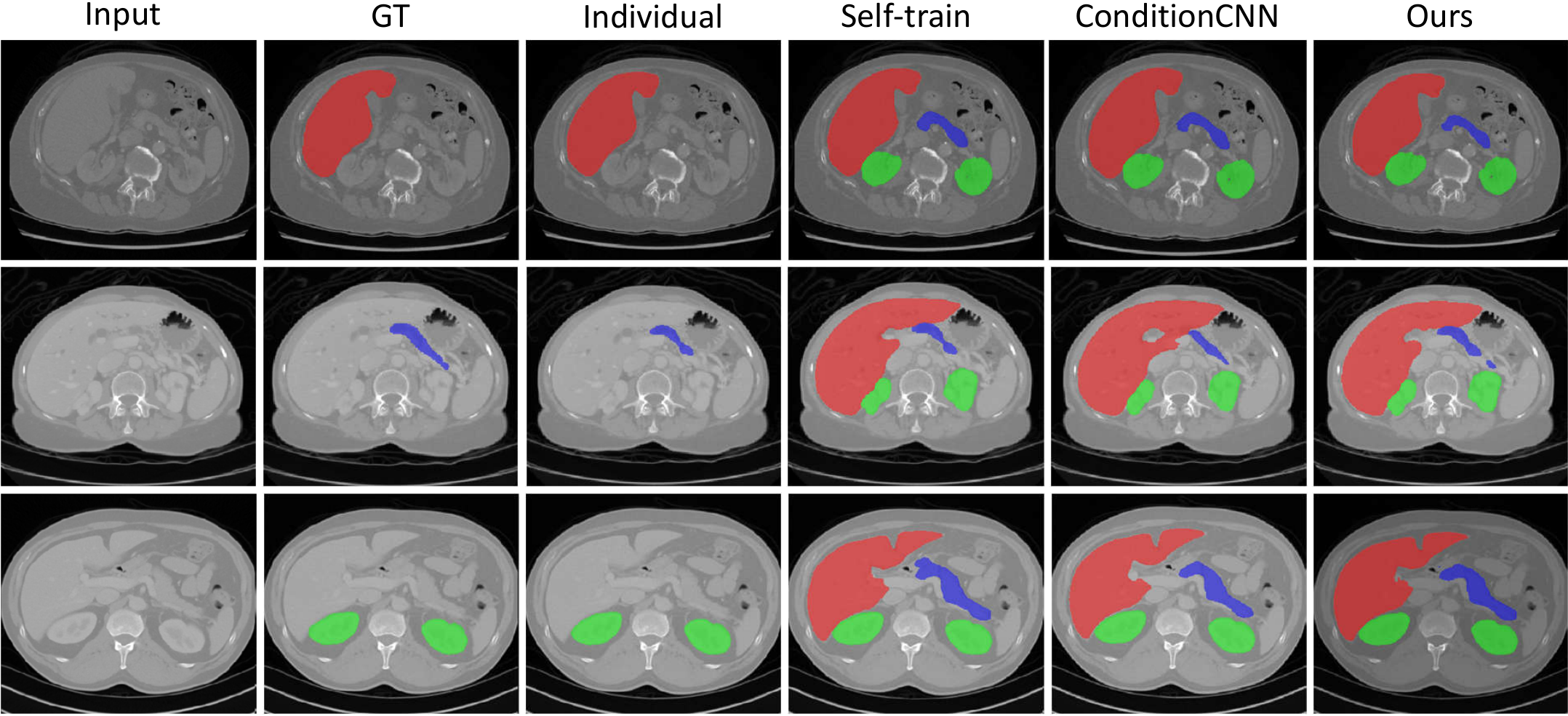}}}
    \caption{Qualitative comparison of different methods. Top to bottom: a LiTS dataset image with liver annotation, a Pancreas dataset image with pancreas annotation, and a KiTS dataset image with kidney annotation.}
    \label{qualitative}
    \end{center}
\end{figure}

{\bf Comparison with state-of-the-art.} We compare our method with state-of-the-art conditionCNN \cite{dmitriev2019learning}, which targets at the same task as our work. Since their full dataset is not publicly available and their code is not open-sourced, we re-implement their method using the above mentioned datasets and our baseline model, and tune the hyper-parameters to achieve their best performance for a fair comparison. The results are shown in Table \ref{tab:2}. We can see that our method outperforms conditionCNN by a considerate margin (0.5\%). In addition, the inference time of conditionCNN is proportion to the number of organs, which makes it inefficient when handling a large number of organs. Some qualitative results are shown in Fig \ref{qualitative}. Our method shows more superior results with less computational cost, compared with existing methods. 

\subsection{Experiments on MOBA dataset}

To validate the generalization ability of our method, we also conduct experiments on MOBA dataset, which is more challenging with eight target organs. Since MOBA has multi-organ annotations, we can train a multi-organ segmentation model directly for comparison. The results are shown in Table \ref{tab:3}. Our method obtains a significant performance gain of 1.19\% compared with the baseline ``individual'' model (82.41\% to 83.60\%). Similarly, a large improvement can be observed for those organs with smaller size and irregular shape, such as esophagus (66.07\% to 72.25\%) and duodenum (58.16\% to 61.63\%), which demonstrate the effectiveness and robustness of our framework. Interestingly, our method even outperforms the fully-supervised results (denoted as ``combine''). We speculate that it might result from  the MOBA has more organs to segment and there is severe class imbalance among organs, and our framework can alleviate the imbalance problem by the proposed online-generated soft pseudo labels. We further calculate Hausdorff Distance in Table \ref{tab:3}. Our method shows much better HD than baseline ``individual'' and self-training strategy, bring gains of 3.94 and 2.63 respectively.

Additionally, 
conditionCNN \cite{dmitriev2019learning} fails to achieve high performance on MOBA. The accuracy drops dramatically, especially for those organs with smaller sizes and irregular shapes. We suspect it's because conditionCNN cannot handle too many organs with high variation by simply incorporating the conditional information into a CNN.


\section{Conclusion}
We propose to co-train weight-averaged models for achieving unified multi-organ segmentation from few-organ datasets. Two networks are collaboratively trained to supervise each other via consistency training. The weight-averaged models are utilized to produce more reliable soft labels for mitigating label noise. Additionally, a region mask is developed to selectively apply the consistent constraint on the regions requiring collaborative teaching. Experiments on four public datasets show that our framework can better utilize the few-organ data and achieves superior performance on multiple public datasets with less computational cost.

\subsubsection{Acknowledgements.} This work is supported in part by the General Research Fund through the Research Grants Council of Hong Kong under Grants CUHK14208417, CUHK14239816, CUHK14207319, in part by the Hong Kong Innovation and Technology Support Programme (No. ITS/312/18FX), in part by the National Natural Science Foundation of China (No. 81871508;  No. 61773246), in part by the Taishan Scholar Program of Shandong Province of China (No. TSHW201502038), in part by the Major Program of Shandong Province Natural Science Foundation (ZR2019ZD04, No. ZR2018ZB0419).

%
%
%
\bibliographystyle{splncs04}
\bibliography{refs}

\end{document}